\documentclass[lettersize,journal]{IEEEtran}
\usepackage{amsmath,amsfonts}
\usepackage{algorithmic}
\usepackage{algorithm}
\usepackage[table,xcdraw]{xcolor}
\usepackage{array}
\usepackage{makecell} 
\usepackage[caption=false,font=normalsize,labelfont=sf,textfont=sf]{subfig}
\usepackage{textcomp}
\usepackage{booktabs,tabulary}
\usepackage{siunitx}
\usepackage{stfloats}
\usepackage{url}
\usepackage[para,online,flushleft]{threeparttable}
\usepackage{multirow}
\usepackage{mathrsfs} \usepackage{amssymb}
\usepackage{verbatim}
\usepackage{graphicx}
\usepackage{color}
\usepackage{cite}
\usepackage{hyperref}
\usepackage[hyphenbreaks]{breakurl}

\begin{document}

\title{Progressive Multi-task Anti-Noise Learning and Distilling Frameworks for Fine-grained Vehicle Recognition}

\author{Dichao Liu

\thanks{e-mail: dc.liu@nagoya-u.jp}


}

\maketitle

\begin{abstract}
Fine-grained vehicle recognition (FGVR) is an essential fundamental technology for intelligent transportation systems, but very difficult because of its inherent intra-class variation. Most previous FGVR studies only focus on the intra-class variation caused by different shooting angles, positions, etc., while the intra-class variation caused by image noise has received little attention. This paper proposes a progressive multi-task anti-noise learning (PMAL) framework and a progressive multi-task distilling (PMD) framework to solve the intra-class variation problem in FGVR due to image noise. The PMAL framework achieves high recognition accuracy by treating image denoising as an additional task in image recognition and progressively forcing a model to learn noise invariance. The  PMD framework transfers the knowledge of the PMAL-trained model into the original backbone network, which produces a model with about the same recognition accuracy as the PMAL-trained model, but without any additional overheads over the original backbone network. Combining the two frameworks, we obtain models that significantly exceed previous state-of-the-art methods in recognition accuracy on two widely-used, standard FGVR datasets, namely Stanford Cars, and CompCars, as well as \textcolor{black}{three additional surveillance image-based vehicle-type classification datasets}, namely \textcolor{black}{Beijing Institute of Technology (BIT)-Vehicle,} Vehicle Type Image Data 2 (VTID2), and Vehicle Images Dataset for Make \& Model Recognition (VIDMMR), without any additional {overheads over} the original backbone networks. The source code is available at {\burl{https://github.com/Dichao-Liu/Anti-noise\_FGVR}.}
\end{abstract}

\begin{IEEEkeywords}
— Fine-grained vehicle recognition, intelligent transportation systems, ConvNets, object recognition
\end{IEEEkeywords}

\section{Introduction}
\begin{figure}[t!]
\centering\includegraphics[width=0.9\linewidth]{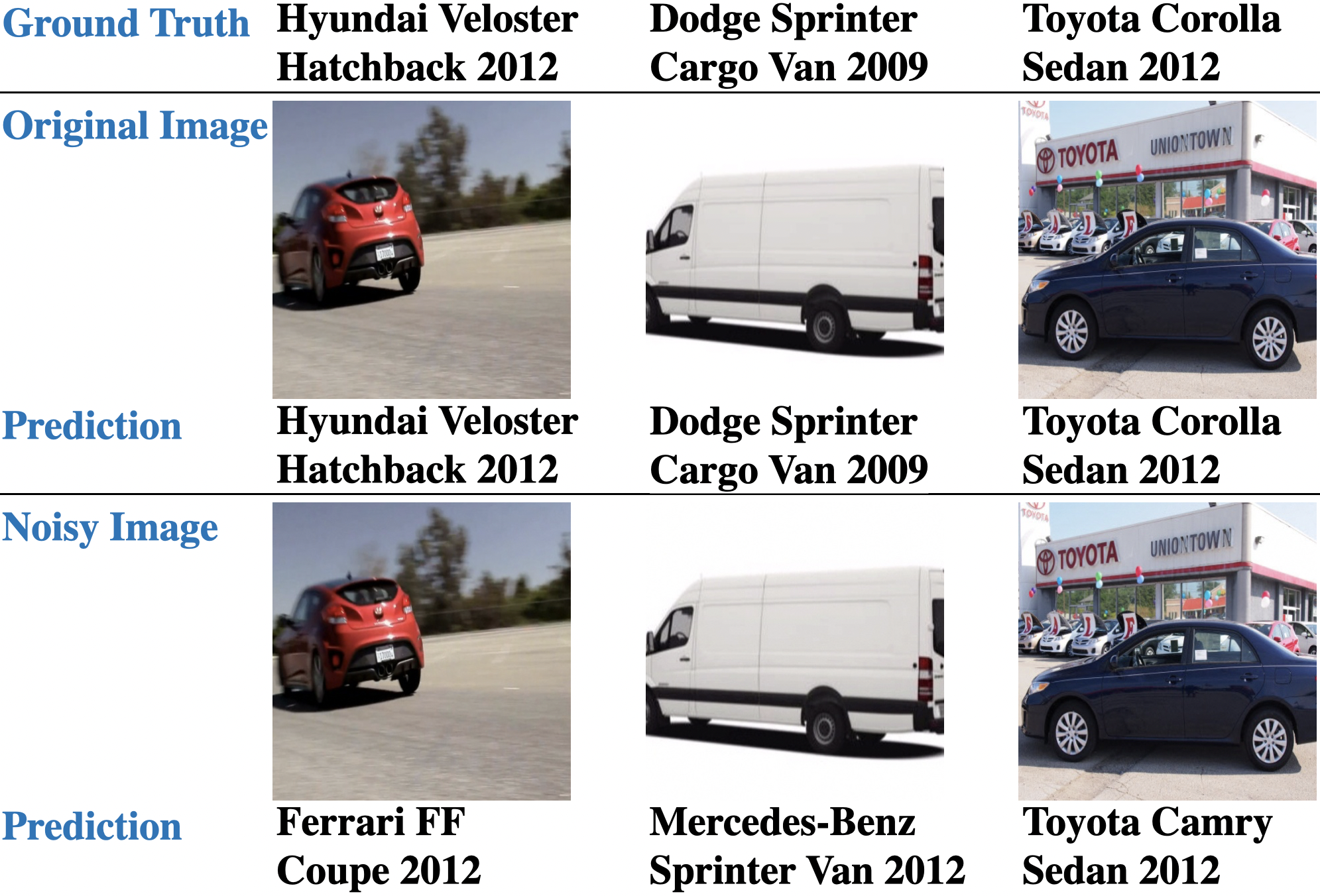}
\caption{Motivation. Convolutional neural networks (CNNs) are very susceptible to noise interruptions in the fine-grained vehicle recognition task. We add a very slight random normal noise (the standard deviation is set as 0.01) to the testing images (normalized to between 0 and 1) from the Stanford Cars dataset~\cite{KrauseStarkDengFei-Fei_3DRR2013} and use a Resnet50~\cite{he2016deep} trained on the original training images of this dataset to recognize the original testing images and noisy testing images, respectively. To the human eye, there is almost no difference between the original image and the noisy image, but the CNN recognizes the noisy images as the wrong models. In real-world intelligent transportation system applications, the obtained vehicle images are commonly affected by image noise. This paper focuses on addressing the intra-class variation caused by image noise.}
\label{fig:noise_problem}
\end{figure}

Fine-grained vehicle recognition (FGVR) refers to the labelling of vehicle images into a number of predefined classes that are typically at the level of {maker}, model, year, e.g. 2007 BMW X5 SUV or 2012 Mercedes-Benz C-Class Sedan~\cite{KrauseStarkDengFei-Fei_3DRR2013}. FGVR has recently attracted a lot of attention from academia and industry due to its wide range of applications, such as identifying specific vehicles with fake license plates in combination with license plate recognition technology~\cite{xiang2019global} and counting the number of specified cars on the highway~\cite{kamkar2016vehicle,liang2015counting}. 

Despite great potentials, FGVR is technically challenging because of the large intra-class variation. Many existing studies try to solve this problem by exploring the ``attention'', which refers to key discriminative information in the given images~\cite{jaderberg2015spatial,hu2017deep,xiang2019global,boukerche2021novel}. For example, some studies locate the attention region and then extract features from the attention region instead of the original image~\cite{jaderberg2015spatial}. Some other studies design various attention modules that can be embedded into deep neural networks and emphasize the features corresponding to discriminative regions~\cite{hu2017deep,xiang2019global,boukerche2021novel}. However, these studies only focus on intra-class variation due to factors in the spatial domain, such as the different shapes or positions of headlights in photographs of cars at different angles, while intra-class variation caused by image noise is largely ignored.\IEEEpubidadjcol

Image noise is the random variations of brightness or color information in images, which can be produced by the process of receiving light as an input signal and outputting the image by the image sensor and circuit of a digital camera. For example, frost or sunlight on the camera may cause random brightness in an image~\cite{buades2005review}. 

In reality, vehicle photos are taken under various natural conditions, and image noise is one of the most important reasons for the intra-class variation of vehicle photographs. Slightly strong noise can result in {unexpected patterns} of the image, while severe noise can make the image difficult for even a human expert to see the vehicle characteristics clearly. However, convolutional neural networks (CNNs), the most commonly used recognition tool nowadays, are very susceptible to noise interruptions, i.e., even small image noise can lead to drastic changes in the output. As shown in Figure~\ref{fig:noise_problem}, image noise that has almost no effect on the human eye can cause a CNN to make mistakes.

In difference to the attention-based approach, we design a novel multi-task learning approach to solve the noise-caused intra-class variation problem, which uses image denoising as an additional task to the recognition task to force the model to learn intra-class invariance. Our approach contains two frameworks. The first is a progressive multi-task anti-noise learning (PMAL) framework, and the second is a progressive multi-task distilling (PMD) framework.

\textbf{PMAL Framework.} The PMAL framework trains the model with the input images added with random normal noise and requires the model to complete two tasks: recognition and image denoising. Specifically, we propose a multi-task learning module, referred to as Denoising-recognition Head (DRH), to solve the tasks. DRH consists of a series of operations, e.g., convolution, pixelshuffle~\cite{huang2009multi}, etc. It takes as input the feature map of a certain CNN layer and outputs a prediction of the input vehicle model and a clean image of the noisy input. During the training, DRH can force the model not only to figure out the category of the vehicle image but also to explore the noise invariance between noisy and clean images. DRH can be developed on top of the feature maps outputted by any layers within a CNN model. {In a CNN, different layers successively abstract low-level information into semantic information from shallow to deep, and the performance of feature extraction in the latter layer depends on the capability of the former layer.} To boost recognition accuracy, we add multiple DRHs to standard CNN models, such as ResNet50, {enabling layers of varying depths to gain noise resistance. However, this manner of proceeding also carries a risk that competition between different tasks may adversely affect recognition accuracy. Specifically, layers of different depths in a CNN capture different levels of information: shallow layers capture low-level concrete information, while deep layers capture high-level semantic information. While DRHs installed in the shallow layers confer noise resistance, they may also compete with DRHs in the deep layers. In such competition, the shallow layer is more likely to dominate the whole training process because it is easier to train, making the deep layers, which tend to contribute more to the recognition accuracy, suffer from unfavorable effects for the recognition task. Therefore, we introduce progressive learning to address this problem. The progressive learning strategy was originally proposed for generative adversarial networks~\cite{karras2018progressive}, where the authors progressively increased the depths of CNN for higher-resolution outputs. In this way, the gradient can be kept to point in a reasonable direction.}

\textbf{PMD Framework.} PMAL yields a powerful FGVR model unaffected by noise-caused intra-class variation. However, more parameters are added compared to the original backbone network (e.g., a plain Resnet50). To address this problem, we propose the PMD framework, a distillation framework designed to match the learning process of the PMAL framework. PMD transfers the knowledge learned by the PMAL-trained model to a plain backbone CNN. Specifically, PMD requires two kinds of tasks: (\romannumeral1) For the layers where DRHs are inserted in the PMAL framework, make the features of the PMAL-trained model in that layer approximated by the features of the plain backbone CNN in the same layer. (\romannumeral2) To make the plain backbone CNN's final prediction score (i.e., softmax distribution) approximate the prediction scores of the model trained by the PMAL framework. These tasks are completed in a progressive manner similar to PMAL to approximate the learning process of PMAL. PMD takes the original input image as the input during training rather than using noisy images as in PMAL. However, PMD receives the knowledge from the PMAL-trained model, so that PMD can make a plain backbone CNN robust against image noise.

Our contributions are summarized as follows:

\begin{itemize}
\item[-] This paper focuses on solving the intra-class variation problem in FGVR caused by image noise, which has received little attention from previous FGVR studies.

\item[-] The proposed PMAL framework includes image denoising as an additional task in image recognition and progressively forces the model to learn noise invariance, which achieves a high recognition accuracy.

\item[-] The proposed PMD framework transfers the knowledge of the PMAL-trained model to the original backbone network, which yields a model with about the same recognition accuracy as the PMAL-trained model but without any additional {computational cost} over the original backbone network.

\item[-] Combining the two frameworks, we obtain models that clearly exceed previous state-of-the-art methods in recognition accuracy on two widely-used standard FGVR datasets, namely Stanford Cars~\cite{KrauseStarkDengFei-Fei_3DRR2013} and CompCars~\cite{yang2015large}, \textcolor{black}{as well as three surveillance image-based vehicle-type classification datasets, namely Beijing Institute of Technology (BIT)-Vehicle~\cite{dong2015vehicle}, Vehicle Type Image Data 2 (VTID2)~\cite{boonsirisumpun2022fast}, and Vehicle Images Dataset for Make \& Model Recognition (VIDMMR)~\cite{ali2022vehicle},} but do not introduce any additional {overheads} over the original backbone networks.

\end{itemize}

\section{Related Studies}

\textbf{Fine-grained Vehicle Recognition.} The mainstream FGVR studies focus on learning discriminative visual clues, i.e., visual attention. The most common attention-based FGVR approaches follow a localization-recognition strategy, where discriminative regions are first localized and then used for recognition~\cite{jaderberg2015spatial}. Doing so can solve the intra-class variation caused by different viewpoints, positions, etc. However, these localization-recognition approaches face some problems, including reliance on laborious additional manual region annotations and the inability to always capture valid regions to the extent that discriminative cues may be overlooked. Many localization-recognition approaches have to crop multiple attention regions to avoid ignoring discriminative cues as much as possible. For example, Jaderberg \textit{et al.}~\cite{jaderberg2015spatial} crop four attention regions and average the prediction scores obtained from the four attention regions as the final prediction score. This method requires a large computational overhead. For each attention region, two networks are required: a localization network and a recognition network. Thus, 8$\times$backbone CNNs (e.g., Resnet50) are required for cropping and recognizing the 4$\times$attention regions.

In order to avoid the drawbacks of localization-recognition strategies, many recent FGVR approaches explore spatial attention information of deep features within CNNs instead of local regions. For example, Hu \textit{et al.}~\cite{hu2017deep} propose a spatially weighted pooling (SWP) layer, which utilizes a learnable mask to estimate the importance of the spatial units of CNN features and pool the features based on the information learned by the mask. Xiang \textit{et al.}~\cite{xiang2019global} propose a global topology constraint network, which uses the constraints of global topological relations to describe the interactions between the CNN features corresponding to local regions and integrates them into the CNN. Boukerche \textit{et al.}~\cite{boukerche2021novel} propose a lightweight recurrent attention unit (LRAU), which can be inserted into standard CNNs to improve the ability to extract discriminative features corresponding to key objects.

These studies have made remarkable contributions to the development of FGVR. However, they focus only on exploring spatial attention to mitigate intra-class variation due to spatial factors (e.g., viewpoint, position in the scene, size of objects, etc.) while ignoring intra-class variation due to image noise. To the best of our knowledge, this paper is the first to focus on intra-class variation due to image noise in FGVR tasks. In addition, the existing FGVR approaches introduce more {overheads} on top of the backbone CNNs. As mentioned above, the localization-recognition strategies require several times the {overheads} of the backbone CNNs on which they are based. Exploring attention information from CNN features is not as burdensome as the localization-recognition strategy. However, additional insertable attention modules still require additional {overheads} beyond the backbone CNNs. In this paper, the PMAL framework also leads to more {overheads}. However, the PMD framework is able to yield a network that is exactly equivalent to the backbone CNN in terms of {computational cost} and network architecture but has an approximate accuracy to the PMAL-trained model.

\textbf{Multi-task Learning.}
Multi-task learning is a training strategy that solves different learning tasks simultaneously to explore the differences or commonalities between tasks. Multi-task learning is inspired by the human behavior of using experience gained in one task to help solve another related task. For example, roller skating experience can help in learning to skate. Multi-task learning is widely used in developing intelligent transportation systems. For example, Zhou \textit{et al.}~\cite{zhou2020unified} propose a unified multi-task learning architecture that jointly solves the object detection task and semantic segmentation task for pedestrian detection.

In our paper, we mainly utilize multi-task learning to force the model to learn the noise-invariant discriminative features. Specifically, in the PMAL framework, the proposed denoising-recognition head (DRH) requires the model to solve the tasks of recognition and image denoising. Image denoising noise refers to recovering the original clean image from a noisy image~\cite{buades2005review}. Traditional image-denoising approaches are often designed based on statistical theory, such as the Gaussian smoothing model~\cite{bruckstein1994gabor}. Recently,  many deep learning-based image-denoising methods have emerged. For example, Huang \textit{et al.} propose a wavelet-inspired invertible network (WINNet)~\cite{huang2022winnet}, which obtains clean images by repeatedly estimating and removing the noise through a carefully designed deep network architecture. In our paper, a new lightweight image-denoising network is designed as part of DRH, which can easily connect the features of any layer of the backbone network and force the whole architecture to be optimized for the image-denoising task. We assume that the image-denoising task helps the model learn noise-invariant discriminative features and better perform the recognition task.

Despite its advantages, multi-tasking learning also has disadvantages. In the training process of multi-task learning, tasks can compete with each other, which may lead to one task dominating the training process. In the PMAL framework, we add DRHs to multiple layers of the backbone networks, forcing the model to learn noise invariance while learning recognition. In deep learning, low-level detail features at shallow layers facilitate image restoration. Higher-level semantic features at deep layers facilitate object recognition. In order to prevent the model from being too dominated by shallow layers, we introduce progressive learning to train the multi-task learning frameworks we designed. Progressive learning refers to training techniques that solve a deep learning problem step by step rather than solving the entire problem at once~\cite{du2020fine,karras2019style,wu2019progressive}. For example, Du \textit{et al.}~\cite{du2020fine} propose a progressive multi-granularity (PMG) training framework for image classification, which effectively fuses the information across different granularities in a step-by-step manner. Liu \textit{et al.}~\cite{LIU2023109550} propose a progressive mutual learning framework, which progressively makes deep and shallow layers learn from each other. In our paper, we progressively train different DRHs to avoid excessive competition between deep and shallow layers when making the model learn noise invariance.

\begin{figure*}[t!]
\centering\includegraphics[width=\linewidth]{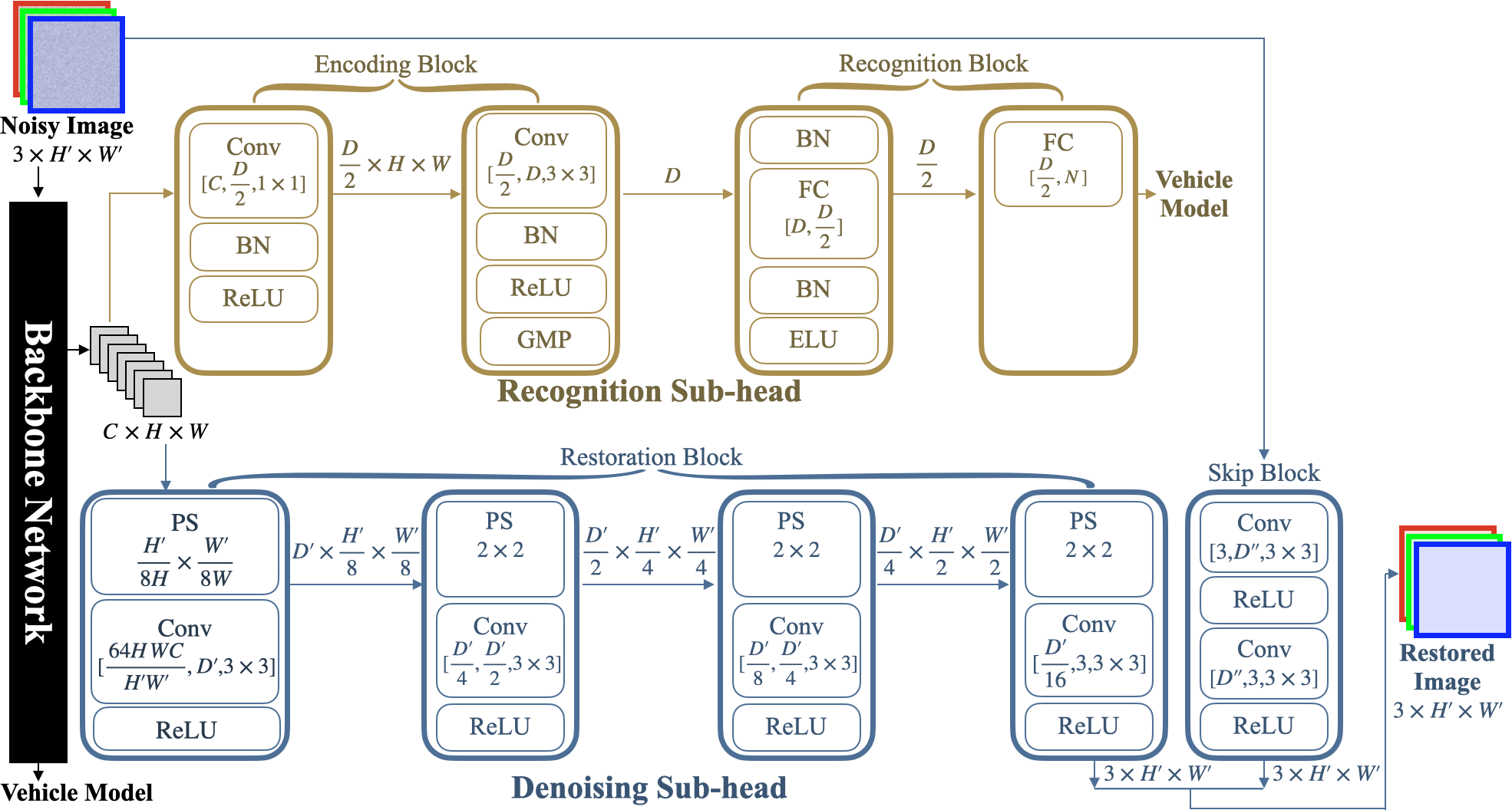}
\caption{Illustration of the architecture of the Denoising-recognition Head (DRH). DRH is based on the intermediate feature from a certain layer of the backbone CNN that is fed a noisy image. DRH consists of a recognition sub-head and a denoising sub-head. The recognition sub-head takes the intermediate feature as input and predicts the vehicle model. The denoising sub-head takes the intermediate feature and the noisy image as input and restores a clean image. In this figure, Conv, ReLU, BN, GMP, FC, ELU, and PS are abbreviations for convolution, rectified linear unit, batch normalization, global maximum pooling, fully-connected, exponential linear unit, and pixelshuffle layers, respectively. Convolution and fully-connected layers are represented by their filter sizes, e.g., $[C,\frac{D}{2},1\times1]$ represents a convolution layer with $C$ input channels, $\frac{D}{2}$ output channels, and a spatial size of $1\times1$. $[D,\frac{D}{2}]$ represents a fully-connected layer with $D$ input neurons and $\frac{D}{2}$ output neurons. In all convolution layers of DRH, the stride is set as 1, and padding is applied to make the spatial size between the input and output constant. Feature maps are represented by ``number of channels''$\times$``height''$\times$``width''. One-dimensional descriptors are represented by their number of neurons. Pixelshuffle layers are represented by their upsampling scale. $D$, $D'$, and $D''$ are manual hyperparameters that control the number of channels in convolution and fully-connected layers.}
\label{fig:drh}
\end{figure*}
\section{Approach}

\subsection{Denoising-recognition Head}

Let $x\in\mathbb{R}^{C\times H\times W}$ denote a feature map from a particular layer of the backbone CNN that is fed a noisy image $I_{\rm noi}\in\mathbb{R}^{3\times H'\times W'}$. $H$, $W$, and $C$ are the height, width, and number of channels of $x$, respectively. $I_{\rm noi}$ is a 3-channel RGB image whose spatial size is $H'\times W'$. Given $x$, DRH gives a prediction of the car model and generates a clean image. DRH consists of two sub-heads: a recognition sub-head $\mathcal{S}_{\rm rec}$ and a denoising sub-head $\mathcal{S}_{\rm den}$.

\textbf{Recognition Sub-head.} The recognition sub-head $\mathcal{S}_{\rm rec}$ {performs} the task of predicting the category of the car model. As shown in Figure~\ref{fig:drh}, $\mathcal{S}_{\rm rec}$ consists of a set of operations, including convolution, pooling, ReLU, etc. Specifically, the function of $\mathcal{S}_{\rm rec}$ can be written as:
\begin{align}
    &p = \mathcal{S}_{\rm rec}(x) = \mathcal{B}_{\rm rec}(d),\\
    &d = \mathcal{B}_{\rm enc} (x),
\end{align}
where $\mathcal{B}_{\rm enc}$ and $\mathcal{B}_{\rm rec}$ are an encoding block and a recognition block, respectively. $\mathcal{B}_{\rm enc}$ encodes $x$ into a one-dimensional descriptor $d\in\mathbb{R}^{D}$ ($D$ is a hyper-parameter), and $\mathcal{B}_{\rm rec}$ yields a prediction score $p\in\mathbb{R}^{N}$ ($N$ is the total number of vehicle model categories) based on $d$. 

As shown in Figure~\ref{fig:drh}, $\mathcal{B}_{\rm enc}$ contains two convolution layers. The filter size of the first convolution layer is $[C,\frac{D}{2},1\times1]$, where $C$, $\frac{D}{2}$, and $1\times1$ denote the input channel number, output channel number, and spatial kernel size, respectively. The filter size of the second convolution layer is $[\frac{D}{2},D,3\times3]$. Batch normalization and ReLU layers are added to both two convolution layers. Then, a global maximum pooling layer pools the feature maps to the one-dimensional descriptor $d$ at the end of $\mathcal{B}_{\rm enc}$. 

At the beginning of the $\mathcal{B}_{\rm rec}$, there is a batch normalization layer to process $d$. The rest of the $\mathcal{B}_{\rm rec}$ consists of two fully-connected (FC) layers. The filter size of the first FC layer is $[D,\frac{D}{2}]$, where $D$ and $\frac{D}{2}$ denote the input channel number and output channel number, respectively. The filter size of the second FC layer is $[\frac{D}{2}, N]$. Batch normalization and exponential linear unit (ELU) layers are added to the first FC layer. The second FC layer acts as a classifier.

\textbf{Denoising Sub-head.} The denoising sub-head $S_{\rm den}$ complements the task of generating a clean image. $S_{\rm den}$ consists of two blocks: a restoration block $\mathcal{B}_{\rm res}$ and a skip block $\mathcal{B}_{\rm ski}$. $\mathcal{B}_{\rm res}$ takes $x$ as input and restores a clean image. $x$ comes from a relatively deep backbone CNN designed for image recognition, thus lacking much of the very shallow information needed for image restoration and possibly suffering from gradient vanishing problems~\cite{tong2017image}. To solve these problems, $\mathcal{B}_{\rm ski}$ takes the noisy input image $I_{\rm noi}$ as input and uses shallow convolution layers to supplement the $\mathcal{B}_{\rm res}$ with low-level features as well as to avoid the gradient vanishing problem interfering with the overall optimization.

Specifically, $\mathcal{B}_{\rm res}$ consists of 4 upsampling modules, $\mathcal{M}_{\rm up}^1$, $\mathcal{M}_{\rm up}^2$, $\mathcal{M}_{\rm up}^3$, and $\mathcal{M}_{\rm up}^4$. $\mathcal{B}_{\rm res}$ can be written as:
\begin{align}
    &I_{\rm res} = \mathcal{B}_{\rm res}(x) = \mathcal{M}_{\rm up}^4(\mathcal{M}_{\rm up}^3(\mathcal{M}_{\rm up}^2(\mathcal{M}_{\rm up}^1(x)))),
\end{align}
where $I_{\rm res}\in\mathbb{R}^{3\times H'\times W'}$ denotes the output image given by $\mathcal{B}_{\rm res}$. As shown in Figure~\ref{fig:drh}, each upsampling module consists of a pixelshuffle layer, a convolution layer, and a ReLU layer. Pixelshuffle~\cite{huang2009multi} is an end-to-end trainable upsampling layer that rearranges a tensor's elements so that the tensor's channel dimension is consumed to expand the spatial dimension to the specified scale. The upsampling scale of the pixelshuffle layer of $\mathcal{M}_{\rm up}^1$ is set to $\frac{H'}{8H}\times\frac{W'}{8W}$, and the upsampling scale of all other pixelshuffle layers is set to $2\times2$. In this way, $\mathcal{M}_{\rm up}^1$ upsamples the spatial size of $x$ from $H\times W$ to $\frac{H'}{8}\times\frac{W'}{8}$, and then $\mathcal{M}_{\rm up}^2$, $\mathcal{M}_{\rm up}^3$, and $\mathcal{M}_{\rm up}^4$ gradually upsample the outputs to $H'\times W'$. The convolution layers in the upsampling modules control the number of channels of the output features with a hyperparameter $D'$. The specific filter size of the convolution layers can be checked in Figure~\ref{fig:drh}.

The function of $\mathcal{B}_{\rm ski}$ can be written as:
\begin{align}
    &I_{\rm ski} = \mathcal{B}_{\rm ski}(I_{\rm noi}),
\end{align}
where $I_{\rm ski}\in\mathbb{R}^{3\times H'\times W'}$ denotes the shallow features to supplement $I_{\rm res}$. $\mathcal{B}_{\rm ski}$ contains two convolution layers. The number of channels of the two convolution layers is defined by a hyperparameter $D''$, and the specific filter size can be checked in Figure~\ref{fig:drh}. 

The denoised image produced by $S_{\rm den}$ is given as:
\begin{align}
    &I_{\rm den} = \mathcal{S}_{\rm den}(x, I_{\rm noi}) = I_{\rm res} + I_{\rm ski}.
\end{align}

\textbf{Loss Function.} The loss function of DRH is the sum of three losses, the first two of which are the softmax loss between $p$ and the ground-truth categorical label, and the mean squared error (MSE) loss between $I_{\rm den}$ and the original input image:
\begin{align}
    &L_{\rm rec} = {\rm Softmax}(p, g),\\
    &L_{\rm den}^{\rm mse} = {\rm MSE}(I_{\rm den}, I),
\end{align}
where $g$ denotes the ground-truth categorical label, and $I\in\mathbb{R}^{3\times H'\times W'}$ denotes the original input image. During the training, $L_{\rm rec}$ and $L_{\rm den}^{\rm mse}$ can force the model to learn to recognize cars and reinforce its understanding of noise invariance at the same time.

MSE loss ensures that $I_{\rm den}$ approximates $I$, but its view of every pixel of the entire image as equally important may make the whole network under-appreciate the visual information important for recognition. Therefore, we re-input the obtained $I_{\rm den}$ into the backbone CNN and let $S_{\rm rec}$ recognize the intermediate features obtained from $I_{\rm den}$. The recognition result is noted as $p_{\rm den}$. Then, the third loss is defined as:
\begin{equation}
    L_{\rm den}^{\rm softmax} = {\rm Softmax}(p_{\rm den}, g).
\end{equation}

The loss function of DRH is defined as:
\begin{equation}
    L_{\rm drh} = L_{\rm rec} + L_{\rm den}^{\rm mse} + L_{\rm den}^{\rm softmax}.
\end{equation}

\subsection{Progressive Multi-task Anti-noise Learning Framework}
\begin{algorithm}[t!]
\caption{Progressive Multi-task Anti-noise Learning}
\label{alg:PMAL}
\begin{algorithmic}[1] 
\FOR{epoch $\in$ [1, num\_of\_epoch]}
\FOR{($I$, $g$) in $\mathcal{V}$}
\FOR{$k$ $\in$ [1, $K$]}
\STATE $I_{\rm noi}^k = {\rm Noise\_Generator}(I)$
\STATE $x^k$ = $\mathcal{C}(I_{\rm noi}^k)$
\STATE $p^k$ = $\mathcal{S}_{\rm rec}^k(x^k)$
\STATE $I_{\rm den}$ = $\mathcal{S}_{\rm den}^k({I_{\rm noi}^k},x^k)$
\STATE $x^k_{\rm den}$ = $\mathcal{C}(I_{\rm den})$
\STATE $p^k_{\rm den}$ = $\mathcal{S}_{\rm rec}^k(x^k_{\rm den})$

\STATE $L^k_{\rm drh}$ = ${\rm Softmax}(p^k,g)+ {\rm MSE}(I_{\rm den},I)+ {\rm Softmax}(p^k_{\rm den},g)$ 
\STATE \textbf{BACKPROP}($L_{\rm drh}^k$)
\ENDFOR
\STATE $I_{\rm noi}^{\mathcal{C}} = {\rm Noise\_Generator}(I)$
\STATE$p^{\mathcal{C}}$ = $\mathcal{C}(I_{\rm noi}^{\mathcal{C}})$
\STATE$L_{\mathcal{C}} = {\rm Softmax}(p^{\mathcal{C}},g)$
\STATE $\textbf{BACKPROP}$($L_{\mathcal{C}}$)
\ENDFOR
\ENDFOR
\end{algorithmic}
\end{algorithm}

\begin{algorithm}[t!]
\caption{Progressive Multi-task Distilling}
\label{alg:PMD}
\begin{algorithmic}[1] 
\FOR{epoch $\in$ [1, num\_of\_epoch]}
\FOR{($I$, $g$) in $\mathcal{V}$}
\FOR{$k$ $\in$ [1, $K$]}
\STATE $x^k_t$ = $\mathcal{T}(I)$
\STATE $x^k_s$ = $\mathcal{C}_s(I)$
\STATE $L^k_{\rm feature}$ = $\alpha$${\rm MSE}$($x^k_s$,$x^k_t$)
\STATE \textbf{BACKPROP}($L^k_{\rm feature}$)
\ENDFOR
\STATE $\{p^1_t$, $p^2_t$, ..., $p^k_t$, ..., $p^K_t\}$, $p^\mathcal{C}_t$ = $\mathcal{T}(I)$
\STATE $p_{s}^{\mathcal{C}} = \mathcal{C}_s(I)$
\STATE$L_{\rm score} = \sum_{k=1}^K {\rm MSE}(p_{s}^{\mathcal{C}},p^{k}_t) + {\rm MSE}(p_{s}^{\mathcal{C}},p^{\mathcal{C}}_t) + {\rm Softmax}(p_{s}^{\mathcal{C}},g)$
\STATE $\textbf{BACKPROP}$($L_{\rm score}$)
\ENDFOR
\ENDFOR
\end{algorithmic}
\end{algorithm}

\begin{figure}[t!]
\centering\includegraphics[width=0.9\linewidth]{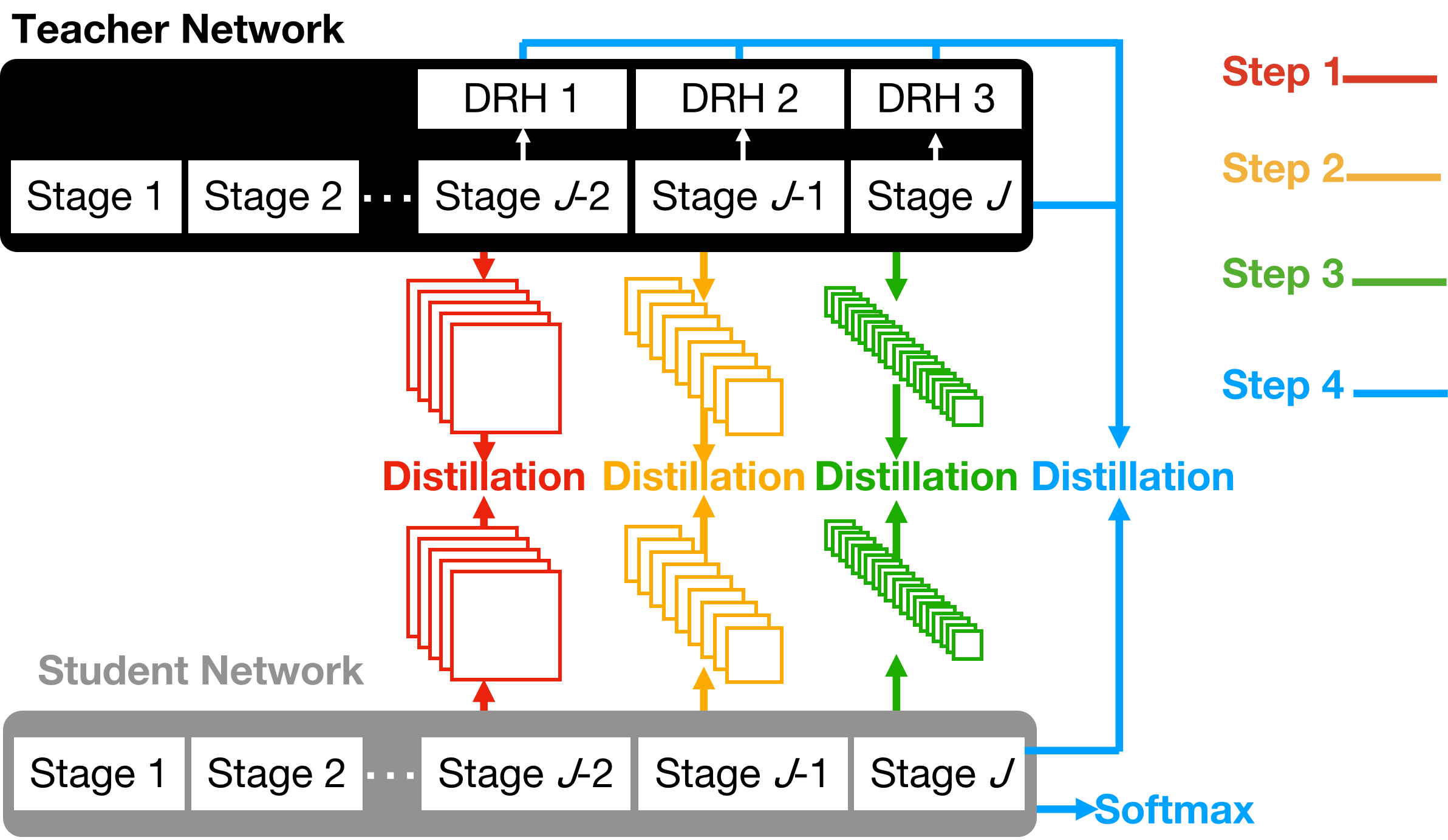}
\caption{{Illustration of the  progressive multi-task distilling framework. Both the teacher and student networks have \textit{J} stages, and three DRHs are installed in the teacher network. Different training steps are illustrated in different colors.}}
\label{fig:dist}
\end{figure}

Let $\mathcal{C}$ be a backbone CNN for recognition, such as ResNet50. In the common case, given an input image, $\mathcal{C}$ outputs a prediction of a vehicle model $p^{\mathcal{C}}$. In this paper, in order to improve the recognition accuracy, we add a total of $K$ DRHs to the plain backbone CNN. Therefore, in training, we need to minimize the losses of $K$ DRHs in addition to the loss between $p^{\mathcal{C}}$ and $g$. 

In a backbone network, layers of different depths tend to learn different levels of information. Shallow layers learn concrete and detailed information, while deep layers learn abstract and semantic information~\cite{zeiler2014visualizing}. In recognition tasks, deep features tend to become dominant during the training, while in low-level vision tasks like image restoration, shallow features tend to be dominant unless sophisticated deep-shallow balancing mechanisms are added~\cite{ren2020scga}. For this paper, recognition is the main task, while image restoration is a secondary task to help recognition. So we must avoid image restoration leading to too much dominance of shallow features. For this purpose, we minimize each loss step by step from shallow to deep rather than minimizing them simultaneously. 

Specifically, as shown in Algorithm~\ref{alg:PMAL}, let $\mathcal{V}$ denote a vehicle image dataset consisting of pairs of images and ground-truth labels, and let $\{x^1$, $x^2$, ..., $x^k$, ..., $x^K\}$ be $K$ intermediate feature maps from shallow to deep layers of $\mathcal{C}$. $\{\mathcal{S}_{\rm rec}^1$, $\mathcal{S}_{\rm rec}^2$, ..., $\mathcal{S}_{\rm rec}^k$, ..., $\mathcal{S}_{\rm rec}^K\}$ and $\{\mathcal{S}_{\rm den}^1$, $\mathcal{S}_{\rm den}^2$, ..., $\mathcal{S}_{\rm den}^k$, ..., $\mathcal{S}_{\rm den}^K\}$ are recognition sub-heads and denoising sub-heads based on $\{x^1$, $x^2$, ..., $x^k$, ..., $x^K\}$, respectively. During the training, the model is optimized in $K+1$ steps in each iteration. At the beginning of each optimization, the noise generator (denoted as ${\rm Noise\_Generator}$) adds the input image with noise. {${\rm Noise\_Generator}$ generates white Gaussian noise with a standard normal distribution defined by a specified standard deviation $\sigma$, and adds the generated noise to the original image, which is normalized to be between 0 and 1.} In the first $K$ steps, we progressively optimize the layers in $\mathcal{C}$ that generate intermediate features $\{x^1$, $x^2$, ..., $x^k$, ..., $x^K\}$ together with the $K$ DRHs based on these intermediate features. In the $K+1$ step, we optimize the recognition prediction given by the whole backbone network of $\mathcal{C}$ (i.e., $p^{\mathcal{C}}$). For the inference, we input the original image to the trained model and get the $K+1$ prediction scores at once. The final prediction is given as:
\begin{align}
    p^{\rm final} = \frac{1}{K+1}(\sum_{k=1}^K p_{\rm ori}^k + p^{\mathcal{C}}_{\rm ori}),
\end{align}
where $\{p^1_{\rm ori}$, $p^2_{\rm ori}$, ..., $p^k_{\rm ori}$, ..., $p^K_{\rm ori}\}$ and $p^{\mathcal{C}}_{\rm ori}$ are the prediction scores obtained with the original input.

\subsection{Progressive Multi-task Distilling Framework}

PMAL framework can improve the recognition accuracy over $\mathcal{C}$, but it also adds additional overheads. For real-world applications, the efficiency of the inference phase is more important than the efficiency during training.
PMAL-trained model does not require denoising sub-heads for inference and only needs to infer once for each input. Therefore, the overhead necessary in the inference phase of the PMAL-trained model is less than training. However, we must admit that there are still additional overheads (i.e., recognition sub-heads) in the PMAL-trained model compared to the original $\mathcal{C}$. Therefore, we propose the PMD framework to obtain a model exactly the same as $\mathcal{C}$ but with higher recognition accuracy.

PMD framework is based on knowledge distillation (KD), which refers to transferring knowledge from a large model (teacher model) to a smaller one (student model). {As shown in Figure~\ref{fig:dist},} PMD framework treats the PMAL-trained model $\mathcal{T}$ as the teacher model and a plain backbone CNN $\mathcal{C}_s$ as the student model. {That is, the student network has the same structure as the backbone on which the teacher network is based. For example, if the teacher network is built on ResNet50 using PMAL, then the student network is a plain ResNet50. Therefore, the teacher and student networks have the same number of stages.} {For training,} both teacher and student models are fed with original input images (rather than noisy images) as input. The student model is required to complement multiple tasks: (\romannumeral1) take $\{x^1_t$, $x^2_t$, ..., $x^k_t$, ..., $x^K_t\}$ from the teacher model as targets and let its intermediate feature maps $\{x^1_s$, $x^2_s$, ..., $x^k_s$, ..., $x^K_s\}$ approximate these targets; (\romannumeral2) make its prediction score $p^\mathcal{C}_{s}$ approximate the teacher model's prediction scores $\{p^1_t$, $p^2_t$, ..., $p^k_t$, ..., $p^K_t\}$ and $p^{\mathcal{C}}_t$; (\romannumeral3) predict the categorical label $g$ of the given input image. 

To coordinate with the learning process of PMAL, we optimize the student model in $K+1$ steps in each iteration of the training of PMD. As shown in Algorithm~\ref{alg:PMD}, In the first $K$ steps, we progressively optimize the approximation of the student model to the teacher model at the intermediate feature level{, i.e. the above-mentioned task~(\romannumeral1)}. In the last one, we optimize the approximation of the student model to the teacher model at the prediction score level{, i.e. the above-mentioned tasks~(\romannumeral2) and (\romannumeral3)}. In Algorithm~\ref{alg:PMD}, $\alpha$ is a hyperparameter controlling the magnitude of the losses in the first $K$ steps. 

\subsection{Sharpness-Aware Minimization}
Breaking down tasks based on layers of different depths into different steps can ensure that the purpose of recognition holistically dominates the learning process. However, doing so may lead to suboptimal results for the optimization of the entire model since each specific step of the optimization may tend to be suboptimal due to several reasons: (\romannumeral1)~the noise added to the input image during PMAL may cause disturbances; (\romannumeral2)~the introduced MSE loss, although it can significantly improve the capture of noise invariance by the model, may tend to be dominated by outliers~\cite{dieguez2020variational,rifai2020evaluation}. These factors reduce the upper limit of our approach for recognition accuracy improvement to some extent. 

To solve the above problems, we introduce Sharpness-Aware Minimization (SAM)~\cite{DBLP:conf/iclr/ForetKMN21} in optimizing each step of PMAL and PMD. SAM is an optimization strategy that simultaneously minimizes loss value and loss sharpness. Specifically, let $w$ be the parameters of the model to be optimized and $\mathcal{L}^k_\mathcal{V}(w)$ be a certain loss ($\mathcal{L}$ can be any of the above-mentioned losses) obtained with $w$ on dataset $\mathcal{V}$ in the $k_{\rm th}$ step ($k\in$$\{1$, $2$, ..., $K$, $K+1\}$). 

In the typical case, the optimization of the model is defined as seeking the $w$ that can yield the minimal loss on dataset $\mathcal{V}$: 
\begin{align}
    \min_w~\mathcal{L}^k_\mathcal{V}(w)+\lambda||w||_2^2, \label{equ:StandardOptimization}
\end{align}
where $\lambda$ is a hyperparameter, and $||w||_2^2$ is a standard L2 regularization term. Instead of Equation~(\ref{equ:StandardOptimization}), SAM seeks parameters that lie in neighborhoods having uniformly low loss:
\begin{align}
    &\min_w~{\rm SAM}(w)+\lambda||w||_2^2,\label{equ:SAM1}\\
    &{\rm where}~~{\rm SAM}(w) = \max_{||\epsilon||_2\leq\rho}~\mathcal{L}^k_\mathcal{V}(w+\epsilon).\label{equ:SAM2}
\end{align}

In Equation~(\ref{equ:SAM2}), $\rho>0$ is a hyperparameter that defines the range of parameter neighborhoods we need to consider when seeking the most optimal parameter. Instead of finding parameters simply with a low training loss, SAM pursues parameters whose entire neighborhoods have uniformly low training loss. In this way, SAM avoids model convergence to a sharp minimum (i.e., a parameter that yields low loss but its neighbor parameters yield high loss), and thus prevents the optimization of each step in our proposed frameworks from converging to a suboptimal result~\cite{DBLP:conf/iclr/ForetKMN21,DBLP:conf/iclr/JiangNMKB20,DBLP:conf/uai/DziugaiteR17}.

\section{Experiments}
\subsection{Datasets and Implementation Details}\label{sec:settings}

\begin{figure}[b!]
\centering
\subfloat[The first DRH]{\includegraphics[width=.5\linewidth]{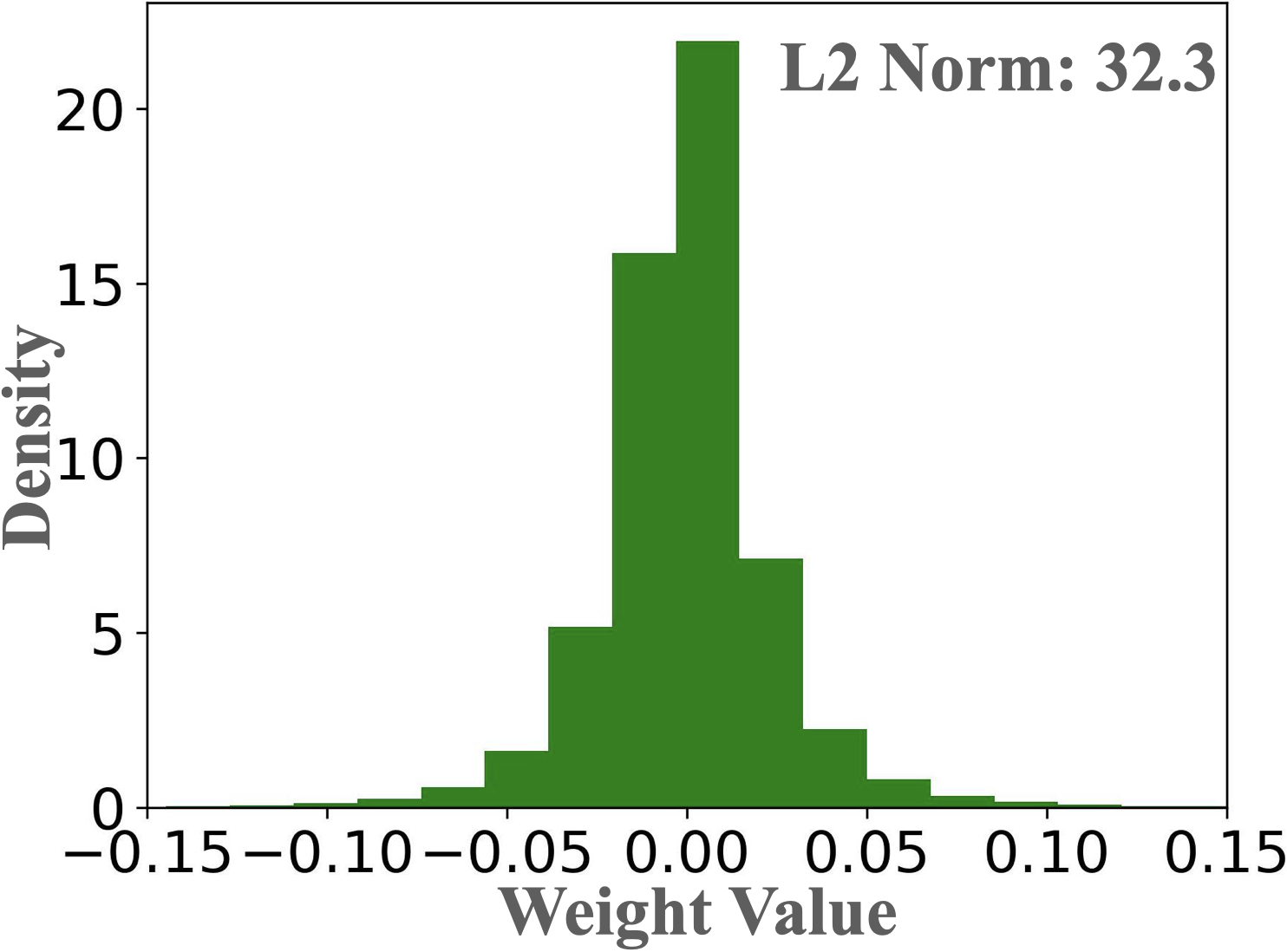}}
\subfloat[The second DRH]{\includegraphics[width=.5\linewidth]{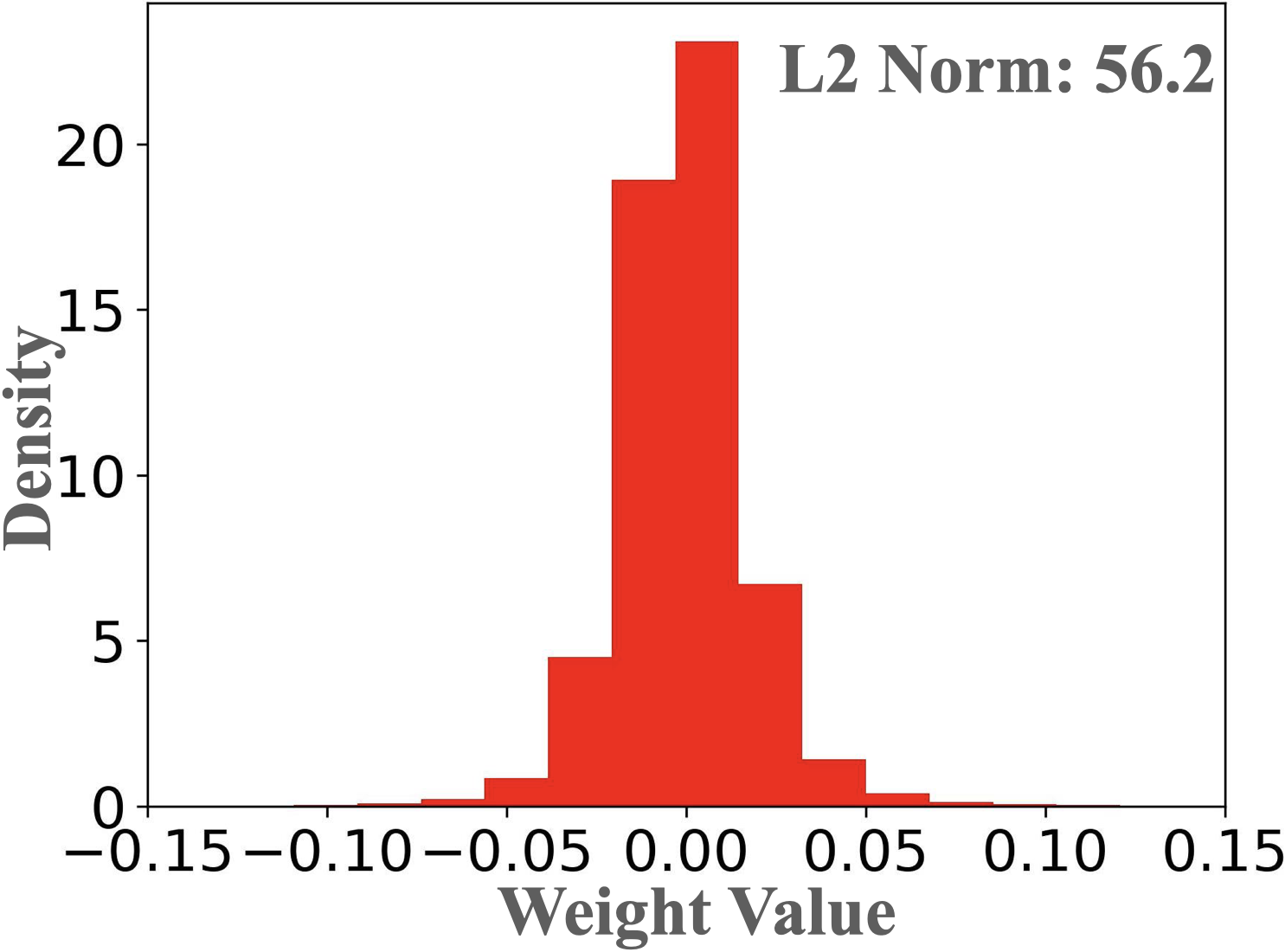}}\hfill
\subfloat[The third DRH]{\includegraphics[width=.5\linewidth]{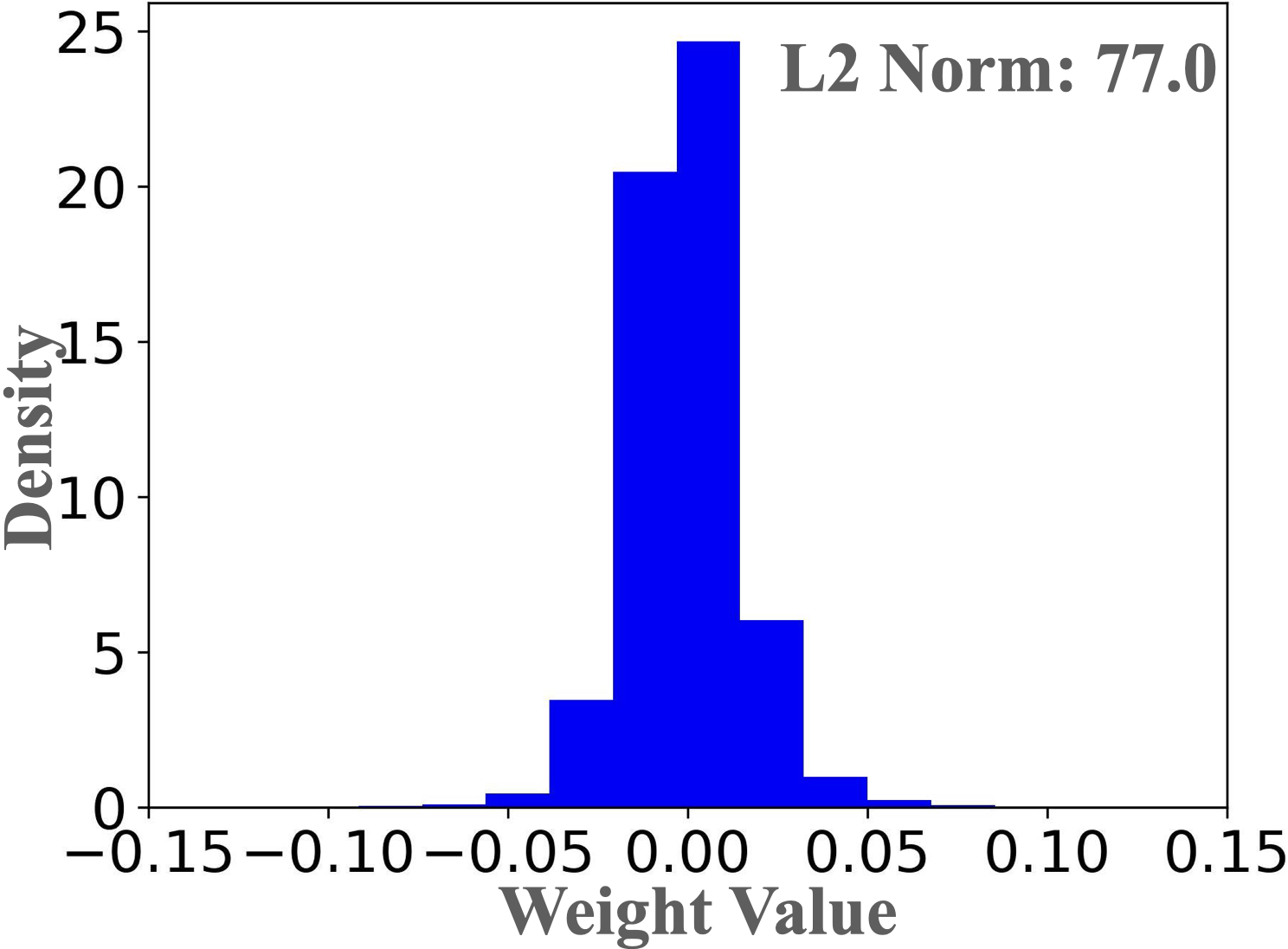}}\hfill
\caption{{In the experiment of this work, we construct three DRHs from shallow to deep layers. This figure illustrates the histograms of the weights associated with the first, second, and third DRHs, respectively. The backbone is a Resnet50 pre-trained on ImageNet~\cite{deng2009imagenet}. The distribution of the three weight histograms is similar, and their L2 norm is in the same order of magnitude. Thus we use the same $\rho$ for the different DRHs in the experiments.}}\label{fig:hist}
\end{figure}

\textbf{Datasets.}~The experiments in this section mianly involve two standard, widely used, and very competitive benchmarks, namely Stanford Cars~\cite{KrauseStarkDengFei-Fei_3DRR2013} and CompCars~\cite{yang2015large}. \textcolor{black}{As two standard datasets in the field of FGVR, Stanford Cars, and CompCars have been widely used by prior studies~\cite{xiang2019global,gao2020channel,luo2020learning, wang2020weakly,boukerche2021novel,zhou2020look,du2020fine,tanveer2021fine,touvron2021grafit,he2021transfg}. Our utilization of the two datasets also follows the prior studies.} 

Stanford Cars is an image dataset with photos of 196 car models, consisting of 8,144 training images and 8,041 testing images. CompCars is a very large dataset containing 431 vehicle models. This dataset has 16,016 training images and 14,939 testing images. Note that although bounding boxes are available with these datasets, the proposed frameworks do not use this extra information. 

\textcolor{black}{In addition, since the number of existing standard vehicle model recognition datasets is limited, in order to further demonstrate the superiority of our method in terms of accuracy, we use three surveillance image-based vehicle-type classification datasets (surveillance images are video frames selected from surveillance videos by the dataset authors), i.e.,  Beijing Institute of Technology (BIT)-Vehicle~\cite{dong2015vehicle}, Vehicle Type Image Data 2 (VTID2)~\cite{boonsirisumpun2022fast}, and Vehicle Images Dataset for Make \& Model Recognition (VIDMMR)~\cite{ali2022vehicle}, as additional datasets, on which to compare the accuracy of our method with the previous state-of-the-art methods on these datasets.}

\textcolor{black}{BIT-Vehicle,} \textcolor{black}{VTID2, and VIDMMR focus on the challenges of keeping up with the times, such as high-resolution input images, the impact of real-life situations on surveillance video, etc.} \textcolor{black}{BIT-Vehicle includes 9850 high-resolution vehicle frontal-view images, which were evaluated with 20-fold validation by the authors proposing this dataset. We follow the same validation strategy as the dataset authors~\cite{dong2015vehicle} in this section. BIT-Vehicle is challenging because it contains changes in the illumination condition, the scale, the surface color of vehicles, and the viewpoint.} \textcolor{black}{VTID2 contains 4,356 images of different types of vehicles that were recorded via surveillance systems. The accuracy of this dataset is evaluated with 10-fold cross-validation by default, and we also follow the default evaluation strategy for this dataset in this section. VIDMMR is a high-resolution image dataset containing 3,847 images that are split into 3,096 training images and 751 testing images. VIDMMR takes into consideration the different viewpoints and illumination effects to approximate the real-world conditions to enhance the challenges. } 

\textbf{Implementation Details.} The experiments are implemented with two standard backbone CNNs, ResNet50~\cite{he2016deep} and TResNet-L~\cite{ridnik2021tresnet}. In the PMAL framework, we add 3 DRHs to the backbones, and accordingly, in the PMD framework, we use 3 intermediate feature maps to transfer knowledge (i.e., $K$ is set as 3). ResNet50 and TResNet-L both have 5 stages. We add DRHs to the end of the last 3 stages. Accordingly, the features outputted by the last 3 stages are used to transfer knowledge. We set the input size as 448$\times$448 for Resnet50 following the common settings in previous studies~\cite{luo2020learning,gao2020channel}. The input size of TResNet-L is set as 368$\times$368 following Ridnik \textit{et al.}~\cite{ridnik2021tresnet}, the authors of this CNN. In this section, we denote the PMAL-trained and PMD-trained models like Resnet50-PMAL, Resnet50-PMD, etc. We train the models with an epoch number of 200 and a mini-batch size of 8. For PMD, the student model is fine-tuned in the last 100 epochs using only softmax loss. The learning rate is set as 0.002 with cosine annealing. The hyperparameters $D$, $D'$, $D''$, $\alpha$, $\lambda$, and $\rho$ are set as 1024, 256, 64, 100, $5\times10^{-4}$, and 0.05. {As shown in Figure~\ref{fig:hist}, the weights associated with the three DRHs are in close magnitude, thus we use the same $\rho$ for the different DRHs in the experiments.} The standard deviation ($\sigma$) of the noise generator defaults to 0.05. The experiments are repeated 8 times, and the average results are reported.

\subsection{Ablation Studies}


\begin{table}[t!]
\caption{{Ablation Studies for training strategies on Stanford Cars and CompCars}}\label{tab:aba}
\setlength{\tabcolsep}{0.3mm}
\begin{threeparttable}
\begin{tabular}{lllllll}
\hline
\multicolumn{2}{l}{Stanford Cars}\\
\multirow{2}{*}{(a)~PMAL} &
  \begin{tabular}[c]{@{}l@{}}Original\\ Resnet50\end{tabular} &
  $K$=1 &
  $K$=2 &
  $K$=3 &
  \begin{tabular}[c]{@{}l@{}}$K$=3\\ (Single Step)\end{tabular} &
  \begin{tabular}[c]{@{}l@{}}$K$=3\\ w/ SAM\end{tabular} \\
 & 93.2$\pm$4\% & 93.9$\pm$3\% & 94.9$\pm$2\% & 95.1$\pm$1\% & 94.7$\pm$3\% & 95.4$\pm$1\% \\ 
\multirow{2}{*}{(b)~PMD} &
  \begin{tabular}[c]{@{}l@{}}Baseline\\ KD~\cite{hinton2015distilling}\end{tabular} &
  $K$=1 &
  $K$=2 &
  $K$=3 &
  \begin{tabular}[c]{@{}l@{}}$K$=3\\ (Single Step)\end{tabular} &
  \begin{tabular}[c]{@{}l@{}}$K$=3\\ w/ SAM\end{tabular} \\
 & 93.9$\pm$3\% & 94.2$\pm$3\% & 94.5$\pm$2\% & 95.1$\pm$1\% & 94.9$\pm$2\% & 95.3$\pm$1\% \\ \hline

\multicolumn{2}{l}{CompCars}\\
\multirow{2}{*}{(c)~PMAL} &
  \begin{tabular}[c]{@{}l@{}}Original\\ Resnet50\end{tabular} &
  $K$=1 &
  $K$=2 &
  $K$=3 &
  \begin{tabular}[c]{@{}l@{}}$K$=3\\ (Single Step)\end{tabular} &
  \begin{tabular}[c]{@{}l@{}}$K$=3\\ w/ SAM\end{tabular} \\
 & 96.9$\pm$3\% & 97.5$\pm$3\% & 97.8$\pm$2\% & 98.5$\pm$1\% & 98.0$\pm$2\% & 99.1$\pm$1\% \\ 
\multirow{2}{*}{(d)~PMD} &
  \begin{tabular}[c]{@{}l@{}}Baseline\\ KD~\cite{hinton2015distilling}\end{tabular} &
  $K$=1 &
  $K$=2 &
  $K$=3 &
  \begin{tabular}[c]{@{}l@{}}$K$=3\\ (Single Step)\end{tabular} &
  \begin{tabular}[c]{@{}l@{}}$K$=3\\ w/ SAM\end{tabular} \\
 & 97.3$\pm$3\% & 97.6$\pm$2\% & 97.9$\pm$2\% & 98.6$\pm$1\% & 98.3$\pm$2\% & 99.0$\pm$1\% \\ \hline
\end{tabular}
\begin{tablenotes}
        \small
        \item[*] Unless otherwise indicated, models embedded with DRH(s) are trained in a multi-step manner by default.
      \end{tablenotes}
\end{threeparttable}
\end{table}

\begin{table}[t!]
\centering
\caption{{Ablation Studies for Hyperparameters on Stanford Cars and CompCars}}\label{tab:aba2}
\setlength{\tabcolsep}{5mm}
\begin{tabular}{llll}
\hline
\multicolumn{2}{l}{Stanford Cars}\\
$D$=256  & $D$=512  & \textbf{$D$=1024} & $D$=2048 \\
94.9$\pm$3\% & 95.0$\pm$1\% & 95.4$\pm$1\% & 95.3$\pm$2\% \\ 
$D'$=64  & $D'$=128 & \textbf{$D'$=256} & $D'$=512 \\
94.8$\pm$4\% & 95.1$\pm$2\% & 95.4$\pm$1\% & 95.2$\pm$3\% \\ 
$D''$=16  & $D''$=32  & \textbf{$D''$=64}  & $D''$=128 \\
94.7$\pm$4\% & 94.9$\pm$1\% & 95.4$\pm$1\% & 94.9$\pm$2\% \\ \hline
\multicolumn{2}{l}{CompCars}\\
$D$=256  & $D$=512  & \textbf{$D$=1024} & $D$=2048 \\
98.5$\pm$2\% & 98.7$\pm$1\% & 99.1$\pm$1\% & 98.7$\pm$1\% \\ 
$D'$=64  & $D'$=128 & \textbf{$D'$=256} & $D'$=512 \\
98.4$\pm$2\% & 98.6$\pm$1\% & 99.1$\pm$1\% & 98.8$\pm$3\% \\ 
$D''$=16  & $D''$=32  & \textbf{$D''$=64}  & $D''$=128 \\
98.5$\pm$3\% & 98.6$\pm$1\% & 99.1$\pm$1\% & 98.8$\pm$1\% \\ \hline
\end{tabular}
\end{table}

We conduct ablation studies for the PMAL and PMD frameworks to evaluate the effectiveness of their design and settings. Resnet50 is used as the backbone network, and Stanford Cars \textcolor{black}{and CompCars} are employed as the datasets in the experiments in this subsection. {In the tables in this subsection, we show the 95\% confidence intervals for the results of the eight experiments in addition to the average accuracy.}

PMAL adds three DRHs optimized by SAM to the original backbone network. To verify the effectiveness of this design, we first train an original Resnet50, then gradually increase the number of DRHs from shallow to deep on top of the original Resnet50, and introduce SAM at last. In addition, to verify the usefulness of multi-step training strategy, we perform an experiment that simultaneously minimizes all the losses of a model with three DRHs in a single step. The results of the above experiments are shown in Table~\ref{tab:aba}~(a) \textcolor{black}{and (c)}. \textcolor{black}{On Stanford Casrs,} adding a single DRH to the original Resnet50 improves accuracy by 0.7\%. As the number of DRHs increases to 3, the accuracy gradually improves, and the introduction of SAM can further improve recognition accuracy. The Resnet50-PMAL finally obtains an accuracy of 95.4\%, which is 2.2\% higher than the accuracy of the original Resnet50. The above experimental results illustrate the effectiveness of the design of PMAL. In addition, the accuracy of optimizing a three-DRH model in one step is not only lower than that of optimizing a three-DRH model in multiple steps, but even lower than that of optimizing a two-DRH model in multiple steps. This experimental result illustrates the effectiveness of the multi-step optimization strategy. \textcolor{black}{On CompCars, similar tendencies were observed. The Resnet50-PMAL with three DRHs obtains the best accuracy of 99.1\%, which is 2.2\% higher than the accuracy of the original Resnet50. } 

In the ablation studies for PMD, we use a plain Resnet50 as the student model and Resnet50-PMAL as the teacher model. As a baseline for comparison, we use only $L_{\rm score}$ of Algorithm~\ref{alg:PMD} (line 11) to train the student model to approach the prediction scores of the Resnet50-PMAL and to predict the ground-truth label, which can be considered a variant of HKD (Hinton's KD)~\cite{hinton2015distilling}, a widely used baseline in previous research~\cite{park2019relational,peng2019correlation}. Afterward, we gradually increase the number of intermediate features used to deliver teacher guidance (i.e., $K$) in the distillation and, at last, introduce SAM. The results are shown in Table~\ref{tab:aba}~(b) \textcolor{black}{and (d)}. \textcolor{black}{On Stanford Casrs,} the accuracy of the student model obtained with $K=1$ is 0.3\% higher than that of the student model trained with baseline KD. The accuracy of the student model gradually improves as $K$ increases to 3 from shallow to deep, and the introduction of SAM can further improve the accuracy. The proposed PMD framework ends up giving the student model an accuracy of 95.3\%, which is a 1.4\% improvement in accuracy over the baseline, a 2.1\% improvement over the original network, and almost the same accuracy as the PMAL-trained model. \textcolor{black}{Similarly, on CompCars, $K=3$ brings the best results 99.0\%, a 2.1\% improvement over the original network and a 1.7\% improvement over the baseline KD.} These results demonstrate the effectiveness of PMD in improving recognition accuracy. In addition, the accuracy of the student model trained by PMD with $K=3$ decreases slightly when multi-step optimization is replaced by single-step optimization, indicating that multi-step optimization also has a positive effect in the PMD framework.

{TABLE~\ref{tab:aba2} shows the results of PMAL using different hyperparameters. We can see that (\romannumeral1) the accuracy of the PMAL-trained model is significantly higher than that of the original network in Table~\ref{tab:aba}~(a), regardless of the hyperparameters used, and (\romannumeral2) the hyperparameters given in Subsection~\ref{sec:settings} work best.}

\subsection{Comparison of the Progressive Multi-task Distilling Framework with Other State-of-the-art Distillation Methods}

\begin{table*}[t!]
\centering
\caption{Comparison of Progressive Multi-task Distilling with Other State-of-the-art Distillation Methods on Stanford Cars \textcolor{black}{and CompCars}}\label{tab:dist}
\setlength{\tabcolsep}{0.2mm}
\begin{tabular}{lllllllllllllllll}
\hline
  SP~\cite{tung2019similarity}&
  Logits~\cite{ba2014deep}&
  Sobolev~\cite{czarnecki2017sobolev} &
  PKT~\cite{passalis2018learning}&
  FSP~\cite{yim2017gift}&
  CC~\cite{peng2019correlation}&
  IRG~\cite{liu2019knowledge}&
  AT~\cite{komodakis2017paying}&
  AB~\cite{heo2019knowledge} &
  LWM~\cite{dhar2019learning}&
  RKD~\cite{park2019relational}&
  Fitnet~\cite{romero2015polytechnique}&
  FPD~\cite{wang2022fpd}&
  TSKD~\cite{xu2023teacher}&
  MFKD~\cite{tan2023knowledge}&
  \textbf{PMD} \\
  \multicolumn{2}{l}{Stanford Cars}\\
\multicolumn{1}{l}{93.8\%} &
  \multicolumn{1}{l}{93.9\%} &
  \multicolumn{1}{l}{93.9\%} &
  \multicolumn{1}{l}{93.9\%} &
  \multicolumn{1}{l}{94.0\%} &
  \multicolumn{1}{l}{94.0\%} &
  \multicolumn{1}{l}{94.0\%} &
  \multicolumn{1}{l}{94.1\%} &
  \multicolumn{1}{l}{94.1\%} &
  \multicolumn{1}{l}{94.1\%} &
  \multicolumn{1}{l}{94.2\%} &
  \multicolumn{1}{l}{94.6\%} &
  \multicolumn{1}{l}{94.5\%} &
  \multicolumn{1}{l}{94.4\%} &
  \multicolumn{1}{l}{94.6\%} &
  \multicolumn{1}{l}{\textbf{95.3\%}} \\ 
  \multicolumn{2}{l}{CompCars}\\
  \multicolumn{1}{l}{97.3\%} &
  \multicolumn{1}{l}{97.3\%} &
  \multicolumn{1}{l}{97.5\%} &
  \multicolumn{1}{l}{97.6\%} &
  \multicolumn{1}{l}{97.5\%} &
  \multicolumn{1}{l}{97.4\%} &
  \multicolumn{1}{l}{97.6\%} &
  \multicolumn{1}{l}{97.7\%} &
  \multicolumn{1}{l}{97.4\%} &
  \multicolumn{1}{l}{97.3\%} &
  \multicolumn{1}{l}{97.2\%} &
  \multicolumn{1}{l}{97.2\%} &
  \multicolumn{1}{l}{98.1\%} &
  \multicolumn{1}{l}{98.1\%} &
  \multicolumn{1}{l}{98.2\%} &
  \multicolumn{1}{l}{\textbf{99.0\%}} \\
  
  \hline
\end{tabular}
\end{table*}

The PMD framework is designed following the learning process of the PMAL framework. Doing so allows the student network to effectively acquire the knowledge of the PMAL-trained model as a teacher network. This subsection shows the superiority of PMD to other state-of-the-art KD approaches~\cite{tung2019similarity,ba2014deep,czarnecki2017sobolev,passalis2018learning,yim2017gift,peng2019correlation,liu2019knowledge,komodakis2017paying,heo2019knowledge,dhar2019learning,park2019relational,romero2015polytechnique,wang2022fpd,xu2023teacher,tan2023knowledge} in the case of using the PMAL-trained model as a teacher network. Concretely, we use Resnet50-PMAL as the teacher model and a plain Resnet50 as the student model. Then, we compare the recognition accuracy of the student model when different KD methods are used to transfer knowledge from Resnet50-PMAL to the student model. The result is shown in Table~\ref{tab:dist}.

KD methods can be roughly classified into two types. The first type of KD approach focuses on making the features of the student model point-to-point to approximate the features of the teacher model~\cite{ba2014deep,romero2015polytechnique,yim2017gift,czarnecki2017sobolev,dhar2019learning,komodakis2017paying,heo2019knowledge,wang2022fpd,xu2023teacher,tan2023knowledge}. For example, Romero \textit{et al.}~\cite{romero2015polytechnique} use the intermediate features learned by the teacher as hints to improve the training process and final performance of the student. Czarnecki \textit{et al.}~\cite{czarnecki2017sobolev} propose using the derivatives generated during training to guide the student model. Some KD methods focus on selecting important features first and then using the selected features for point-to-point distillation~\cite{dhar2019learning,komodakis2017paying,heo2019knowledge}. The PMD proposed in this paper can be classified as a point-to-point KD method.

The second type of KD approach focuses on making the distribution or correlation of the features of the student model approximate that of the teacher model~\cite{park2019relational,peng2019correlation,liu2019knowledge,tung2019similarity,passalis2018learning}. For example, Park \textit{et al.}~\cite{park2019relational} propose distance-wise and angle-wise distillation losses that penalize the differences between the feature relationship between the student and teacher. 

As shown in Table~\ref{tab:dist}, \textcolor{black}{on Stanford Cars,} the accuracy of the student model trained with PMD is 0.7\%--1.5\% higher than that of the student model trained with other methods. \textcolor{black}{On CompCars, the accuracy of the student model trained with PMD is 0.8\%--1.7\% higher than that of the student model trained with other methods.} This demonstrates that PMD transfers knowledge more effectively than other KD methods when using the PMAL-trained model as a teacher.

\begin{figure*}[t!]
\centering\includegraphics[width=\linewidth]{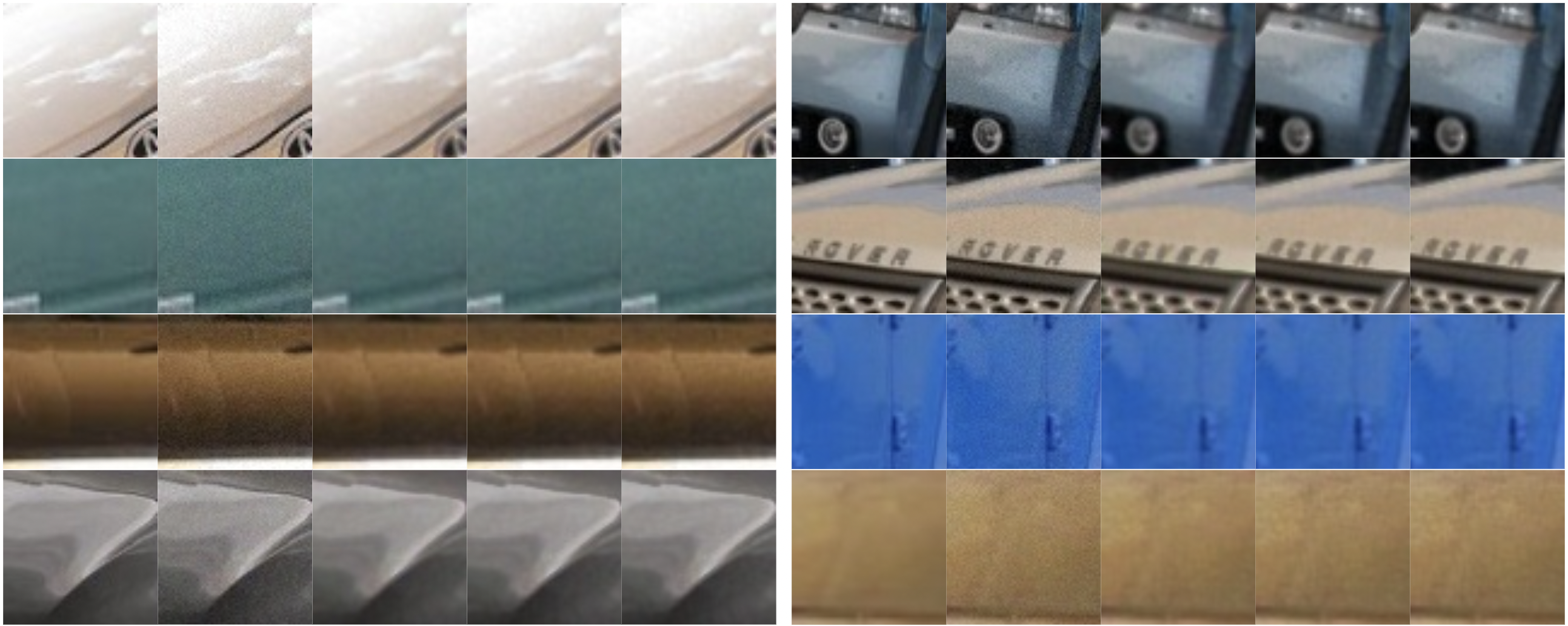}
\caption{Denoised results. Each group of five images, from left to right, shows the original image, the noisy image, and the noise reduction results derived from $S_{\rm den}^1$, $S_{\rm den}^2$, and $S_{\rm den}^3$. Given noisy images, $S_{\rm den}^1$, $S_{\rm den}^2$, and $S_{\rm den}^3$ can generate smooth and noise-free images.}
\label{fig:denoise_results_samples}
\end{figure*}

\begin{figure}[t!]
\centering\includegraphics[width=\linewidth]{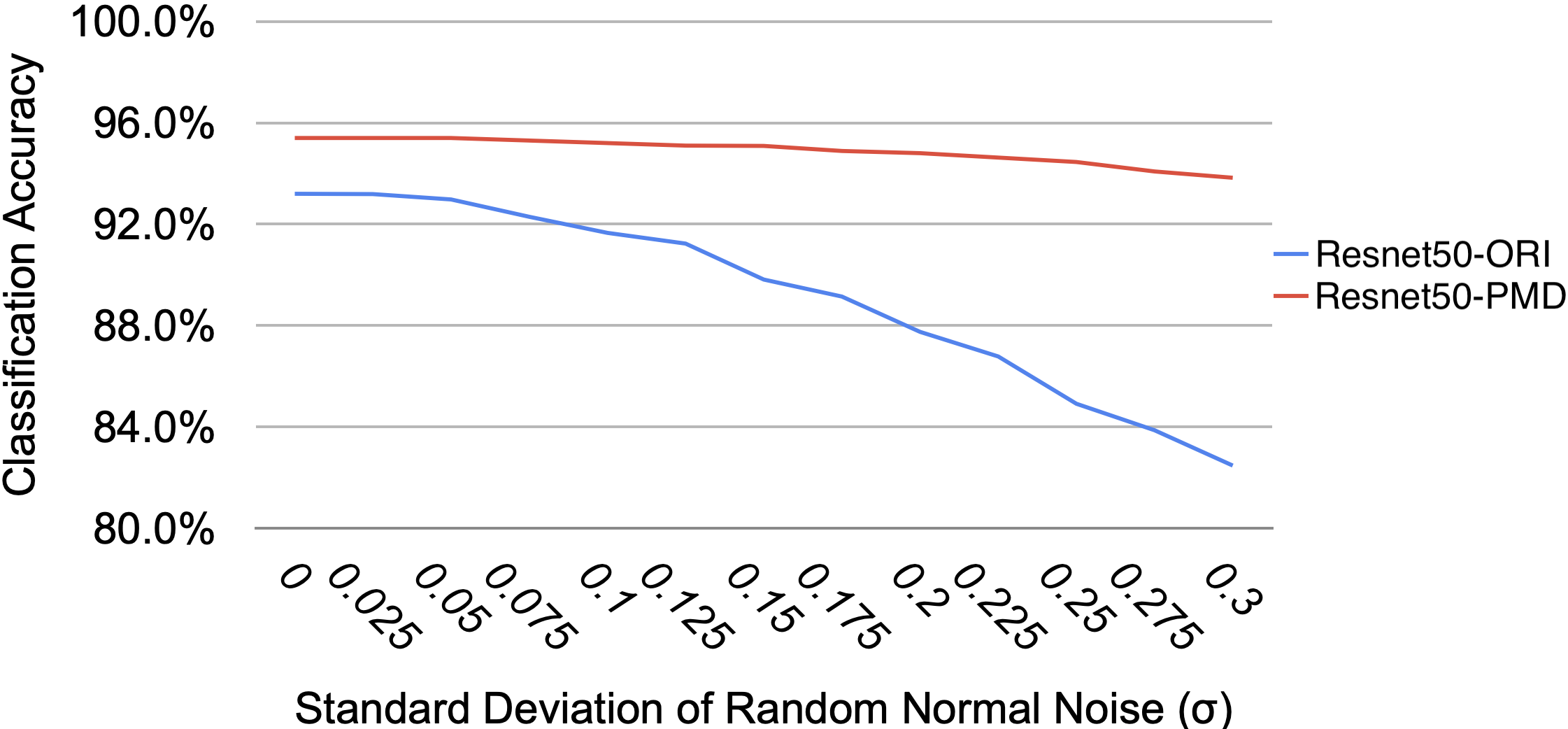}
\caption{Recognition accuracy with increasing noise power (i.e., the standard deviation of random normal noise~$\sigma$).}
\label{fig:denoise_results_accuracy}
\end{figure}

\subsection{Effectiveness in Capturing Noise Invariance}
To explain the effectiveness of the proposed method, in this subsection, we show the impact of the proposed method on the ability to capture the noise invariance. 

First, we show the effect of image denoising by the 3 DRHs inserted into the backbone network in the PMAL framework. As shown in Figure~\ref{fig:denoise_results_samples}, given a noisy image, all 3 DRHs can give a relatively smooth image with the noise removed. This result indicates that the PMAL-trained model acquires the ability to capture noise invariance~\cite{chatterjee2009clustering,kalampokas2021moment}. Also, referring to Table~\ref{tab:aba}~(a) and (c), it can be seen that the improvement of noise invariance capturing capability brought by PMAL to the model can improve the recognition accuracy of the model.

Second, we quantitatively show that the PMD framework can transfer the noise invariance learned by the PMAL-trained model to the student network and make the student network robust against image noise. Concretely, We first train two models on the training set of the Stanford Cars dataset: (\romannumeral1)~Resnet50-PMD; (\romannumeral2)~an original Resnet50 trained without using the PMD framework, which is denoted as Resnet50-ORI. These two models have exactly the same architecture. Then, we apply random normal noise to the images in the test set of the Stanford Cars dataset and let Resnet50-PMD and Resnet50-ORI recognize the noisy images, respectively. We gradually increase the noise power (i.e., $\sigma$) to observe the recognition accuracy of the two models. As shown in Figure~\ref{fig:denoise_results_accuracy}, the recognition accuracy of Resnet50-ORI decreases sharply with the rise of noise power. In contrast, the decrease in the recognition accuracy of Resnet50-PMD is minimal.

Note that in the PMD framework, the input is the original image without noise applied, and the learning of noise invariance is totally dependent on the guidance of the PMAL-trained model as a teacher. Moreover, in the training of PMAL, $\sigma$ is set to 0.05, while the student model that obtains the knowledge transferred from the PMAL-trained model in the PMD framework still has high recognition accuracy even in the face of noise with $\sigma$ = 0.3. These results demonstrate that our method effectively improves the capture of noise invariance.

\subsection{Comparison to State-of-the-art Fine-grained Vehicle Recognition Methods}

\begin{table}[t!]
\centering
\caption{\mbox{Comparison with state-of-the-art approaches on Stanford Cars}}\label{tab:sota_stanford}
\begin{tabular}{lll}
\hline
Approach                                     & Backbone  & Accuracy \\\hline
SEF (SPL, 2020~\cite{luo2020learning}) & Resnet50                               & 94.0\% \\
GCL (AAAI, 2020~\cite{gao2020channel})       & Resnet50  & 94.0\%   \\
CIN (AAAI, 2020~\cite{wang2020graph})                        & Resnet101           & 94.5\% \\
DF-GMM (CVPR, 2020~\cite{wang2020weakly})    & Resnet50  & 94.8\%   \\
LIO (CVPR, 2020~\cite{zhou2020look})         & Resnet50  & 94.5\%   \\
PMG (ECCV, 2020~\cite{du2020fine})         & Resnet50  & 95.1\%   \\
Resnet101-GTC (T-ITS, 2020~\cite{xiang2019global})          & Resnet101 & 94.0\%   \\
Desnet264-GTC (T-ITS, 2020~\cite{xiang2019global}) & Desnet264& 94.3\%   \\
Grafit (ICCV, 2021~\cite{touvron2021grafit})                 & Efficientnet-B7 & 94.7\% \\
DeiT (ICML, 2021~\cite{touvron2021training}) & DeiT-B    & 93.3\%   \\
NAT (TPAMI, 2021~\cite{lu2021neural})        & NAT-M4    & 92.9\%   \\
AutoFormer (ICCV, 2021~\cite{chen2021autoformer})            & AutoFormer-S                          & 93.4\% \\
TransFG (AAAI, 2022~\cite{he2021transfg})                    & ViT-B\_16  & 94.8\% \\
VGG16-LRAU (T-ITS, 2022~\cite{boukerche2021novel})          & VGG16  & 93.5\%   \\
Resnet50-LRAU (T-ITS, 2022~\cite{boukerche2021novel})          & Resnet50  & 93.9\%   \\
CMAL (Pattern Recognition, 2023~\cite{LIU2023109550})          & TResnet-L  & 97.1\%   \\
 \hline
\textbf{PMAL (Ours)}                        & Resnet50  & 95.4\%   \\
                                             & TResnet-L & 97.3\%   \\
\textbf{PMD (Ours)}                         & Resnet50  & 95.3\%   \\
                                             & TResnet-L & 97.3\%   \\ \hline
\end{tabular}
\end{table}

\begin{table}[t!]
\centering
\caption{\mbox{Comparison with state-of-the-art approaches on CompCars}}\label{tab:sota_comp}
\setlength{\tabcolsep}{1.1mm}
\begin{tabular}{lll}
\hline
Approach                       & Backbone    & Accuracy \\ \hline
Resnet152-CMP (TVT, 2019~\cite{ma2019fine})       & Resnet152   & 97.0\%   \\
Densenet161-CMP (TVT, 2019)    & Densenet161 & 97.9\%   \\
Fine-Tuning DARTS (ICPR, 2020~\cite{tanveer2021fine}) & NAS-Model         & 95.9\%   \\
Resnet101-GTC (T-ITS, 2020~\cite{xiang2019global})    & Resnet101   & 98.3\%   \\
Desnet264-GTC (T-ITS, 2020~\cite{xiang2019global})    & Desnet264   & 98.5\%   \\
Resnet50-LRAU (T-ITS, 2022~\cite{boukerche2021novel})    & Resnet50    & 98.3\%   \\
VGG16-LRAU (T-ITS, 2022~\cite{boukerche2021novel})       & VGG16       & 98.3\%   \\ \hline
\textbf{PMAL (Ours)}           & Resnet50    & 99.1\%   \\
                               & TResnet-L   & 99.0\%   \\
\textbf{PMD (Ours)}            & Resnet50    & 99.0\%   \\
                               & TResnet-L   & 98.9\%   \\ \hline
\end{tabular}
\end{table}

\begin{table}[t!]
\centering
\caption{\mbox{Comparison with state-of-the-art approaches on BIT-Vehicle}}\label{tab:sota_bit}
\setlength{\tabcolsep}{0.8mm}
\begin{tabular}{lll}
\hline
Approach                       & Backbone    & Accuracy \\ \hline
Fixed Feature Extraction (IEEE TVT, 2020~\cite{soon2020semisupervised})    & Resnet50   & 87.1\%   \\
PCN-Softmax (IEEE TVT, 2020~\cite{soon2020semisupervised})    & PCN   & 88.5\%   \\
TC-SF-CNNLS (ICIEA, 2021~\cite{awang2021performance})    & SF-CNNLS   & 90.5\%   \\
Network Mapping (Access, 2022~\cite{lin2022intelligent})    & YOLO-CFNN    & 90.5\%   \\
DFT (Expert Systems, 2022~\cite{zhang2022dft})       & VGG16       & 90.1\%   \\ 
Tri-Training (Expert Systems, 2022~\cite{zhang2022dft})       & VGG16       & 90.8\%   \\ 
IBSA (JIT, 2023~\cite{chen2023vehicle})    & IBSA-CNN    & 87.7\%   \\
\hline
\textbf{PMAL (Ours)}           & Resnet50    & 95.2\%   \\
                               & TResnet-L   & 94.9\%   \\
\textbf{PMD (Ours)}            & Resnet50    & 94.9\%   \\
                               & TResnet-L   & 94.7\%   \\ \hline
\end{tabular}
\end{table}

\begin{table}[t!]
\centering
\caption{{Results of additional experiments on VTID2 and VIDMMR}}\label{tab:additional}
\begin{threeparttable}
\setlength{\tabcolsep}{5mm}
\begin{tabular}{lll}
\hline
                               & VTID2  & VIDMMR  \\ \hline
MobileNets                     & 94.4\% & 73.5\%  \\
Inception V4                   & 92.2\% & -       \\
Inception V3                   & 91.4\% & -       \\
Resnet50                       & 91.2\% & 67.1\%  \\
Resnet152                      & -      & 69.2\%  \\
Inception Resnet V2            & 93.0\% & -       \\
Darknet-19                     & 91.6\% & -       \\
Darknet-53                     & 93.4\% & -       \\
MobileNetV2                    & 93.6\% & -       \\
VGG16                          & 77.5\% & 74.3\%  \\
\hline
\textbf{Resnet50+PMAL (Ours)}  & 97.9\% & 100.0\% \\
\textbf{TResnet-L+PMAL (Ours)} & 97.7\% & 100.0\% \\
\textbf{Resnet50+PMD (Ours)}   & 97.7\% & 100.0\% \\
\textbf{TResnet-L+PMD (Ours)}  & 97.4\% & 100.0\% \\ \hline
\end{tabular}
\begin{tablenotes}
        \small
        \item[*] The accuracy of the prior approaches listed in this table is cited from the official results reported in~\cite{boonsirisumpun2022fast} or~\cite{ali2022vehicle}.
      \end{tablenotes}
\end{threeparttable}
\end{table}

This subsection demonstrates the superiority of the proposed approach to the previous state-of-the-art FGVR method in terms of recognition accuracy. The comparison of the recognition accuracy of the models trained by the proposed frameworks with the previous state-of-the-art FGVR methods on Stanford Cars and CompCars is shown in Tables~\ref{tab:sota_stanford} and \ref{tab:sota_comp}, respectively. Even though previous state-of-the-art methods already possess high and near-saturated accuracy on the two datasets, our method still significantly surpasses them. \textcolor{black}{The comparison results on BIT-Vehicle~\cite{dong2015vehicle} are shown in Table~\ref{tab:sota_bit}. The comparison results on VTID2~\cite{boonsirisumpun2022fast} and VIDMMR~\cite{ali2022vehicle} are shown in Table~\ref{tab:additional}.}

Concretely, on Stanford Cars, TResnet-L-PMD achieves the same accuracy as TResnet-L-PMAL, and Resnet50-PMD is only 0.1\% lower than Resnet50-PMAL. Most previous state-of-the-art methods on this dataset adopt Resnet50 as the backbone, and the proposed Resnet50-PMD and Resnet50-PMAL outperform all the previous approaches using the same backbone. Moreover, the previous approaches introduce additional mechanisms to the original backbones, but the proposed Resnet50-PMD has exactly the same architecture as the original Resnet50. Using TResnet-L as the backbone network, the proposed frameworks deliver higher accuracy than all the previous state-of-the-art methods.

On CompCars, Resnet50-PMD and TResnet-L-PMD are just 0.1\% lower than Resnet50-PMAL and TResnet-L-PMAL, respectively. The networks of the Resnet family are also the most commonly used backbone networks for previous state-of-the-art approaches on this dataset. Using Resnet50 and TResnet-L as backbone networks, the proposed frameworks give similar recognition accuracy, significantly higher than the previous methods on this dataset. 

\textcolor{black}{On the} \textcolor{black}{three} \textcolor{black}{additional datasets, namely} \textcolor{black}{BIT-Vehicle,} \textcolor{black}{VTID2, and VIDMMR, the improvement of our method is very significant compared to the previous approaches.} \textcolor{black}{On BIT-Vehicle, our different networks outperform the previous best method by 4.1\% to 4.4\%.} \textcolor{black}{On VTID2, our networks outperform the previous best method by 3.0\% to 3.5\%, and on VIDMMR, our networks achieve 100\% accuracy. On all three additional datasets, PMD-trained models can achieve the same or similar accuracy as the PMAL-trained models.}

\textcolor{black}{The above results demonstrate the superiority of our method over the prior methods in terms of accuracy. In addition, note that the student network trained under the PMD framework has exactly the same structure as the original backbone network. It means that our approach does not need to introduce all kinds of complex additional mechanisms like the prior approaches to achieve accuracy beyond them, which is another superiority of our approach.}

\section{Conclusion}

This paper focuses on solving the intra-class variation problem caused by image noise in FGVR. The proposed PMAL framework includes image denoising as an additional task in image recognition and forces the model to learn noise invariance in a multi-step manner from shallow to deep layers. The experimental results demonstrate that using image denoising as an additional task can effectively make the model learn noise invariance and improve recognition accuracy. Furthermore, the multi-layer progressive anti-noise training strategy used in the PMAL framework can make the whole model noise-invariant from the extraction of low-level information to the refinement of high-level information. Based on the above advantages, the PMAL framework can give powerful models with high accuracy. In addition, this paper also proposes the PMD framework, which transfers the knowledge of PMAL-trained models to the original backbone networks in a multi-step manner from shallow to deep layers. PMD is designed to be coordinate with the learning process of PMAL. This design is proved to perform better than previous knowledge distillation methods while using the PMAL-trained model as the teacher. PMD can yield models that reach the approximate accuracy of PMAL-trained models but with a computational burden equivalent to that of the original backbone networks. Our approach also exhibits resistance to noise attacks, illustrating that our approach can indeed address the image noise problem that this paper envisions solving. The method proposed in this paper outperforms previous state-of-the-art methods on Stanford Cars, CompCars, \textcolor{black}{BIT-Vehicle}, VTID2, and VIDMMR without introducing additional computational burden on top of the original backbone networks. 


\bibliographystyle{IEEEtran}
\bibliography{egbib}

\vfill

\end{document}